\documentclass[journal]{IEEEtran}
\usepackage{multirow}
\usepackage{booktabs}

\ifCLASSINFOpdf
  \usepackage[pdftex]{graphicx}
  \graphicspath{{../pdf/}{../jpeg/}}
\else
  \usepackage[dvips]{graphicx}
  \usepackage{}
\fi
\usepackage{amsmath}

\begin{document}
%
\title{Generating Multi-scale Maps from Remote Sensing Images via Series Generative Adversarial Networks}
%
%
%

\author{Xu~Chen, Bangguo~Yin, Songqiang~Chen, Haifeng~Li~\IEEEmembership{Member,~IEEE}, and~Tian~Xu
\thanks{\emph{Corresponding author:~Tian~Xu~(email:xutian@whu.edu.cn)}}
\thanks{Xu~Chen, Bangguo~Yin, Songqiang~Chen, and~Tian~Xu are with School of Computer Science, Wuhan University, Wuhan, 430072, China.}
\thanks{Haifeng Li is with School of Geosciences and Info-Physics, Central South University, Changsha, 410083, China.}
}


%
%

\markboth{IEEE Geoscience and Remote Sensing Letters,~Vol.~14, No.~8, August~2015}%
{Shell \MakeLowercase{\textit{et al.}}: Generating Multi-scale Maps from Remote Sensing Images via Series Generative Adversarial Networks}
%



\maketitle

\begin{abstract}
Considering the success of generative adversarial networks (GANs) for image-to-image translation, researchers have attempted to translate remote sensing images (RSIs) to maps (rs2map) through GAN for cartography. However, these studies involved limited scales, which hinders multi-scale map creation. By extending their method, multi-scale RSIs can be trivially translated to multi-scale maps (multi-scale rs2map translation) through scale-wise rs2map models trained for certain scales (parallel strategy). However, this strategy has two theoretical limitations. First, inconsistency between various spatial resolutions of multi-scale RSIs and object generalization on multi-scale maps (RS-m inconsistency) increasingly complicate the extraction of geographical information from RSIs for rs2map models with decreasing scale. Second, as rs2map translation is cross-domain, generators incur high computation costs to transform the RSI pixel distribution to that on maps. Thus, we designed a series strategy of generators for multi-scale rs2map translation to address these limitations. In this strategy, high-resolution RSIs are inputted to an rs2map model to output large-scale maps, which are translated to multi-scale maps through series multi-scale map translation models. The series strategy avoids RS-m inconsistency as inputs are high-resolution large-scale RSIs, and reduces the distribution gap in multi-scale map generation through similar pixel distributions among multi-scale maps. Our experimental results showed better quality multi-scale map generation with the series strategy, as shown by average increases of 11.69\%, 53.78\%, 55.42\%, and 72.34\% in the structural similarity index, edge structural similarity index, intersection over union (road), and intersection over union (water) for data from Mexico City and Tokyo at zoom level 17–13.
\end{abstract}


%
\IEEEpeerreviewmaketitle

\section{Introduction}

\IEEEPARstart{W}{ith} the emergence of deep learning, it is possible to bypass vectorization when translating remote sensing images (RSIs) to maps by employing generative adversarial networks (GANs) \cite{goodfellow_generative_2014}. GAN has been widely applied for image-to-image translation (img2img) which aims to learn mapping between an input image and an output image \cite{zhu_unpaired_2017}. As a typical model for img2img, the Pix2Pix method \cite{isola_image--image_2017} uses L1 distance as a conditional loss in GAN to achieve the goal of img2img, with experimental results indicating mutual translation between map tiles and aerial photos. Inspired by these results, specific trials have been conducted to translate RSIs to maps (\textbf{rs2map translation}) \cite{ganguli_geogan_2019, chen_smapgan_2020, li_mapgan_2020}. Although these trials have achieved some good results with rs2map translation, their experiments only focused on certain map scales. \cite{ganguli_geogan_2019} only used satellite images at a zoom level of 14 (approximately 7.24 meters per pixel), \cite{chen_smapgan_2020} only used RSIs of 2.15 meters per pixel, and \cite{li_mapgan_2020} appeared to only employ images with the same spatial resolution. 

However, the production of multi-scale maps is important for cartography. Traditionally, multi-scale maps were produced using the cartography generalization method. Automated mapping methods offer operators for cartography generalization, such as selection, simplification, and displacement \cite{feng_learning_2019}, which are typically operated on vectorized data. With the aid of deep learning, \cite{ijgi9050338,feng_learning_2019} revealed the potential for generalization of mountain roads and building. 

In this study, we aim to develop an end-to-end method for creating a multi-scale map inspired by previous attempts at rs2map translation (\textbf{multi-scale rs2map translation}).  The trivial strategy employs parallel scale-wise rs2map models \textbf{(r2m generators)} to translate RSIs of various scales to maps of corresponding scales to translate RSIs of various scales to maps of corresponding scales (\textbf{parallel strategy}). However, according to our analysis, this strategy suffers from two theoretical limitations. First, the inconsistency between various RSI spatial resolutions of multi-scale RSI and object generalization (\textbf{RS-m inconsistency}) on multi-scale maps makes it increasingly difficult to extract the required geographical information from RSIs for rs2map models with decreasing scale. Second, as rs2map translation is cross-domain translation, there is a large gap between the pixel distributions of RSIs and maps at the same scale, and r2m generators incur high computational costs at each scale in the parallel strategy (\textbf{multi cross-domain translation}). 

To address these limitations, we design a series strategy for multi-scale rs2map translation. In this strategy, high-resolution RSIs are inputted to an rs2map model to output large-scale maps, which are translated to multi-scale maps through series multi-scale map translation models (\textbf{m2m generators}). The series strategy avoids RS-m inconsistency because the inputs are high-resolution RSIs of a certain large scale. Moreover, the strategy reduces the distribution gap in multi-scale map generation because the pixel distributions between maps of neighboring scales are less than those between RSIs and maps according to the Earth Mover’s Distance (EMD) \cite{rubner_earth_2000}. 

Our experimental results demonstrate that multi-scale maps generated by the series strategy are better quality than those generated by the parallel strategy. This is shown by an average increase of 11.69\%, 53.78\%, 55.42\%, and 72.34\% in the structural similarity index  (SSIM) \cite{wang_image_2004}, the edge structural similarity index (ESSI) \cite{chen_smapgan_2020}, intersection over union (IOU) \cite{jaccard_distribution_1912} (road), and intersection over union (water) for the data of Mexico City and Tokyo at a zoom level of 17 to zoom 13.    

The remainder of this paper proceeds as follows. Section II analyzes the limitations of the parallel strategy and discusses the series strategy proposed to address these limitations. Section III describes the detailed structure of the series and parallel strategies. Section IV presents the experiments and analyzes the results. Section V concludes the study and proposes improvements for future work.

\section{Analysis of Multi-scale Rs2map Translation}
The trivial parallel strategy for multi-scale rs2map translation suffers from two theoretical limitations. First, variations in the spatial resolution of RSIs are not consistent with object generalization on maps. Second, the pixel distributions of RSIs and maps differ substantially in range and shape, leading to high computational costs for r2m generators in the parallel strategy.
\subsection{Limitations of the Parallel Strategy}
Equation \ref{eq1} presents rs2map translation:
\begin{align}
	\label{eq1}
	\mathcal{G}_{r\rightarrow m}(r)=m
\end{align}
where $\mathcal{G}_{m\rightarrow r}$ denotes the r2m generator, $ r $ denotes the RSI, and  $ m $ denotes the generated map. Thus,  \ref{eq3} presents multi-scale rs2map translation:
\begin{align}
	\label{eq3}
	\mathcal{S}_{R_{ms}\rightarrow M_{ms}}(R_{ms})=M_{ms}
\end{align}
where $\mathcal{S}_{R_{ms}\rightarrow M_{ms}}$ denotes the strategy of generators, $R_{ms}$ denotes the set of multi-scale RSIs, and $M_{ms}$ denotes the set of generated multi-scale maps.

The parallel strategy of multi-scale rs2map translation at each scale is shown by  \ref{eq2}:
\begin{align}
	\label{eq2}
	\mathcal{G}_{r_{scale}\rightarrow m_{scale}}(r_{scale})=m_{scale}
\end{align}
where $scale$ denotes a certain scale. Then the parallel strategy involves a set of generators as \ref{eq4}:
\begin{equation}
\begin{aligned}
	\label{eq4}
	\mathcal{S}_{R_{ms}\rightarrow M_{ms}}=(&\mathcal{G}_{r_{scale0}\rightarrow m_{scale0}}, \\&\mathcal{G}_{r_{scale1}\rightarrow m_{scale1}}, \\&\mathcal{G}_{r_{scale2}\rightarrow m_{scale2}}, ...)
\end{aligned}
\end{equation}

The parallel strategy aims to extract the variation rule of multi-scale maps from the variation of multi-scale RSIs, as shown in \ref{eq5}: 
\begin{equation}
	\begin{aligned}
		\label{eq5}
	\mathcal{S}_{R_{ms}\rightarrow M_{ms}}(\mathcal{V}_{R_{ms}})=\mathcal{V}_{M_{ms}}	
	\end{aligned}
\end{equation}
where $\mathcal{V}$ denotes the variation rule of multi-scale RSIs or maps. 

\begin{figure}[htbp]
	\centering
	\includegraphics[scale=0.3]{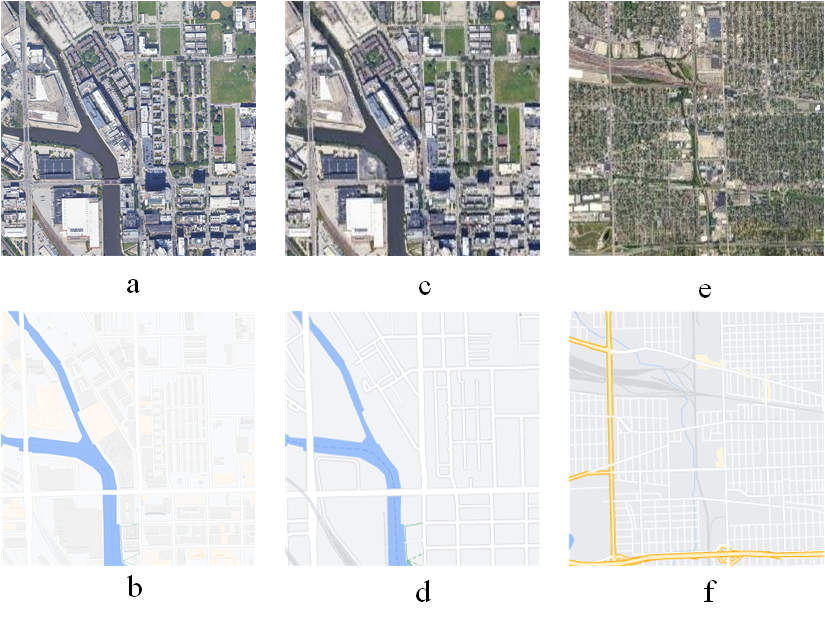}
	\caption{Comparison of multi-scale RSIs and maps from Google Map at a zoom level of (a) and (b) 17 (512×512, 0.6 m/pixel); (c) and (d) 16 (512 × 512, 1.2 m/pixel); and (e) and (f) 13 (512 × 512, 9.5 m/pixel).}
	\label{Figure 1}
\end{figure}

However, this strategy suffers from inconsistency between the variation rules of multi-scale RSIs and maps (\textbf{RS-m inconsistency}). The change of objects on multi-scale maps does not always correspond to that on multi-scale RSIs, as shown in the images taken at different zoom levels Fig. \ref{Figure 1}. The Buildings in Fig. \ref{Figure 1} (a) are still visible in Fig. \ref{Figure 1} (b) and (c); however, the buildings are erased in Fig. \ref{Figure 1} (d).  

Consequently, the signal-to-noise ratio (SNR) of RSIs decreases as the scale and spatial resolution decrease. As details on maps are generalized from small scale to large scale, details of RSIs tend to become noise. In Fig. \ref{Figure 1} (e), the buildings become noisy pixels covering most of the image. However, Fig. \ref{Figure 1} (f) only maintains the road net, which only covers a small part on the RSI. Thus, a decrease in the scale of the RSIs reduces the ability of r2m generators to distinguish and extract effective information in rs2map translation.   
\begin{figure}[htbp]
	\centering
	\includegraphics[scale=0.1]{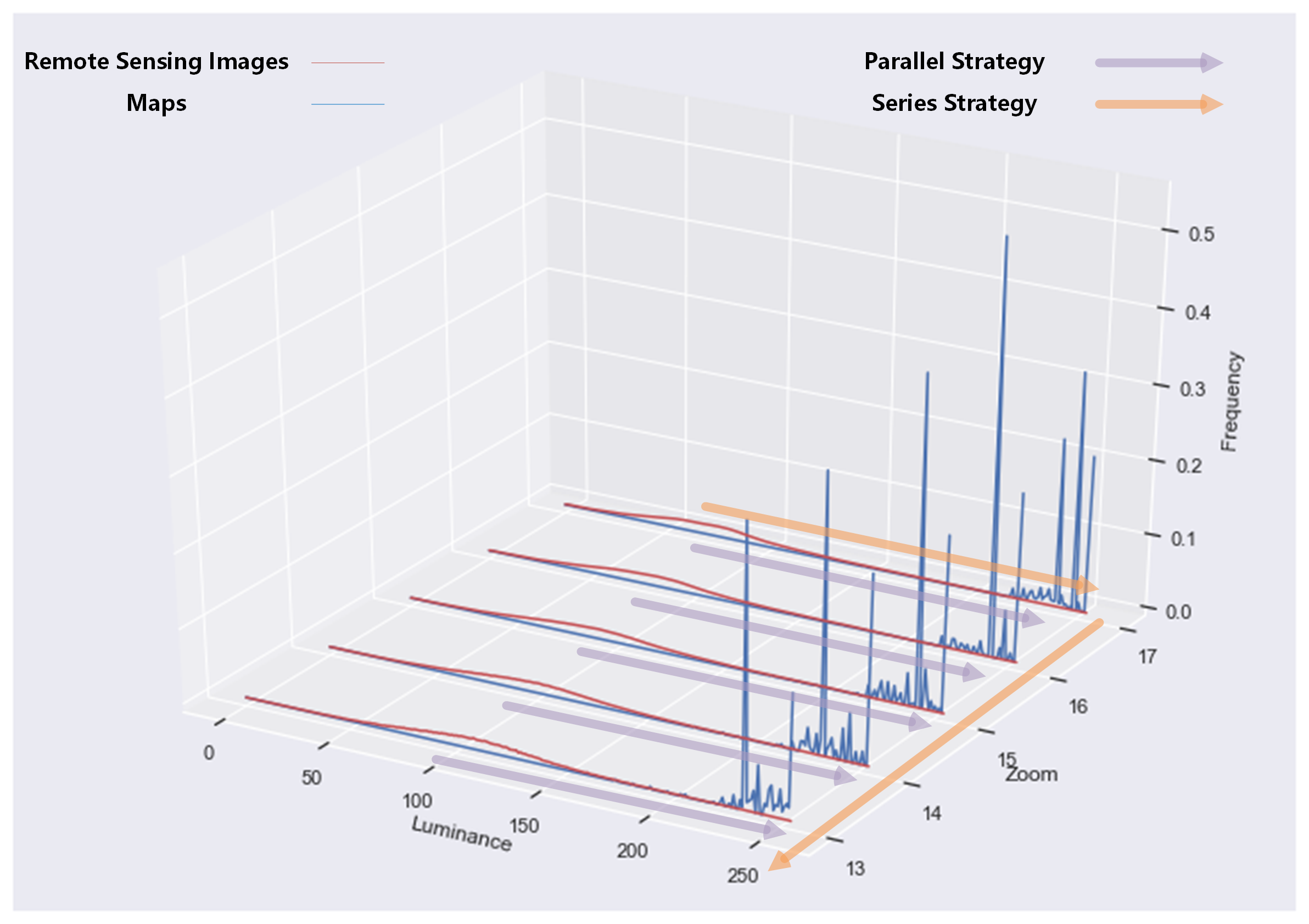}
	\caption{Distribution of pixels on RSIs and maps.}
	\label{Figure 2}
\end{figure}

In addition, rs2map translation is cross-domain translation. RSIs and maps represent two visual expressions of geographic information in which the pixels of RSIs and maps are distributed over relatively different ranges, as shown in Fig. \ref{Figure 2}. Furthermore, the pixel distributions of RSIs and maps have very different shapes, with the distribution of RSIs not as clustered as that of maps.

As the core of GAN is to transform one distribution to another \cite{pmlr-v70-arjovsky17a}, \ref{eq2} can be rewritten as  \ref{eq6}:
\begin{align}
	\label{eq6}
	\mathcal{G}_{r_{scale}\rightarrow m_{scale}}(P_{x\sim R_{scale}})=P_{x\sim M_{scale}}
\end{align}
where $ P_{x\sim R_{scale}} $ and $ P_{x\sim M_{scale}} $ denote distributions of pixels on RSIs and maps at a certain scale. Then, \ref{eq4} can be rewritten as \ref{eq7}:
\begin{equation}
	\begin{aligned}
		\label{eq7}
		\mathcal{S}_{R_{ms}\rightarrow M_{ms}}=(&\mathcal{G}_{P_{x\sim R_{scale0}}\rightarrow P_{x\sim M_{scale0}}}, \\&\mathcal{G}_{P_{x\sim R_{scale1}}\rightarrow P_{x\sim M_{scale1}}}, \\&\mathcal{G}_{P_{x\sim R_{scale2}}\rightarrow P_{x\sim M_{scale2}}}, ...)
	\end{aligned}
\end{equation}

Thus, scale-wise cross-domain translation should be conducted with the parallel strategy, which may lead to complications.
\subsection{Series Strategy: Addressing the Limitations of the Parallel Strategy}
The design of the series strategy enables us to solve the above problems. The series strategy divides multi-scale rs2map translation into two parts: translation from high-resolution RSIs to large-scale maps and translation among multi-scale maps. Thus, the learning variation rule of multi-scale maps is determined by the potential connection among multi-scale maps instead of the variation of multi-scale RSIs.

The series strategy involves in two types of generators: an r2m generator from high-resolution RSIs to large-scale maps ($ \mathcal{G}_{r_{hs}\rightarrow m_{ls}} $) and series scale-wise m2m generators from large-scale to small-scale maps ($ \mathcal{G}_{m_{ls}\rightarrow m_{ss}} $). Equation \ref{eq8} presents this strategy:
\begin{equation}
	\begin{aligned}
		\label{eq8}
		\mathcal{S}_{R_{hs}\rightarrow M_{ms}}=...(&\mathcal{G}_{m_{scale2}\rightarrow m_{scale3}}(\\&\mathcal{G}_{m_{scale1}\rightarrow m_{scale2}}(\\&\mathcal{G}_{m_{scale0}\rightarrow m_{scale1}}(\\&\mathcal{G}_{r_{scale0}\rightarrow m_{scale0}}(R_{hs})))))...
	\end{aligned}
\end{equation}

Consequently, the negative impact of RS-m inconsistency is eliminated. Moreover, a decrease in the SNR of multi-scale RSIs will not disturb the series strategy because only high-resolution RSIs of the largest scale ($ scale0 $) and highest spatial resolution are employed as inputs, which have the highest SNR.

Moreover, only one cross-domain translation is conducted by $ \mathcal{G}_{r_{hs}\rightarrow m_{ls}} $ in the series strategy, and the other m2m generators are approximately intra-domain translation. M2m generators are employed to translate maps of a certain scale to those of a neighbouring smaller scale, with both map scales exhibiting similar pixel distributions, as shown in Fig. \ref{Figure 2}.

\begin{table}[htbp]\scriptsize
	\centering
	\caption{Earth mover's distance of the parallel strategy and series strategy at different zoom levels}
	\begin{tabular}{cccccc}
		\toprule[1pt]
		\textbf{STRATEGY}  & \textbf{17} & \textbf{16} & \textbf{15} & \textbf{14}&\textbf{13} \\
		\midrule
		\textbf{Parallel Strategy} & 0.005782 &0.006339 &0.006066 &0.005909 &0.005336 \\ 
		\textbf{Series Strategy} &  0.005782 &0.001789 &0.000973&0.000611 &0.000700 \\		
		\bottomrule[1pt]
	\end{tabular}%
	\label{table1}%
\end{table}%

The EMD is applied in this study to measure the minimum cost of transformation from one distribution to another. Table \ref{table1} shows the EMD of the parallel and series strategies for multi-scale rs2map translation (zoom level 17 to 13 in Google Maps) at each scale. The results indicate that a much lower cost is required for m2m generators to translate large-scale maps to smaller-scale maps in the series strategy than for rs2map translation at a smaller scale in the parallel strategy.

\section{Methodology}
\subsection{Selection of Generator: Pix2PixHD}
We choose Pix2PixHD \cite{wang_high-resolution_2018} as the generator because it can translate high-resolution images, and the inputs and outputs were 512 × 512 in size as well as our data in experiment. Pix2PixHD employs one generator to generate a low-resolution image, which is then input to another generator to generate a high-resolution image. Thus, with this strategy, the output can balance global and local features from the input.

In addition, Pix2PixHD employs multi-scale discriminators to discriminate the output. The output is 2× and 4× downsampled and discriminated using multi-scale discriminators. The average discrimination result was then used to guide the generators. 

\subsection{The Series Strategy}
\begin{figure}[htbp]
	\centering
	\includegraphics[scale=0.14]{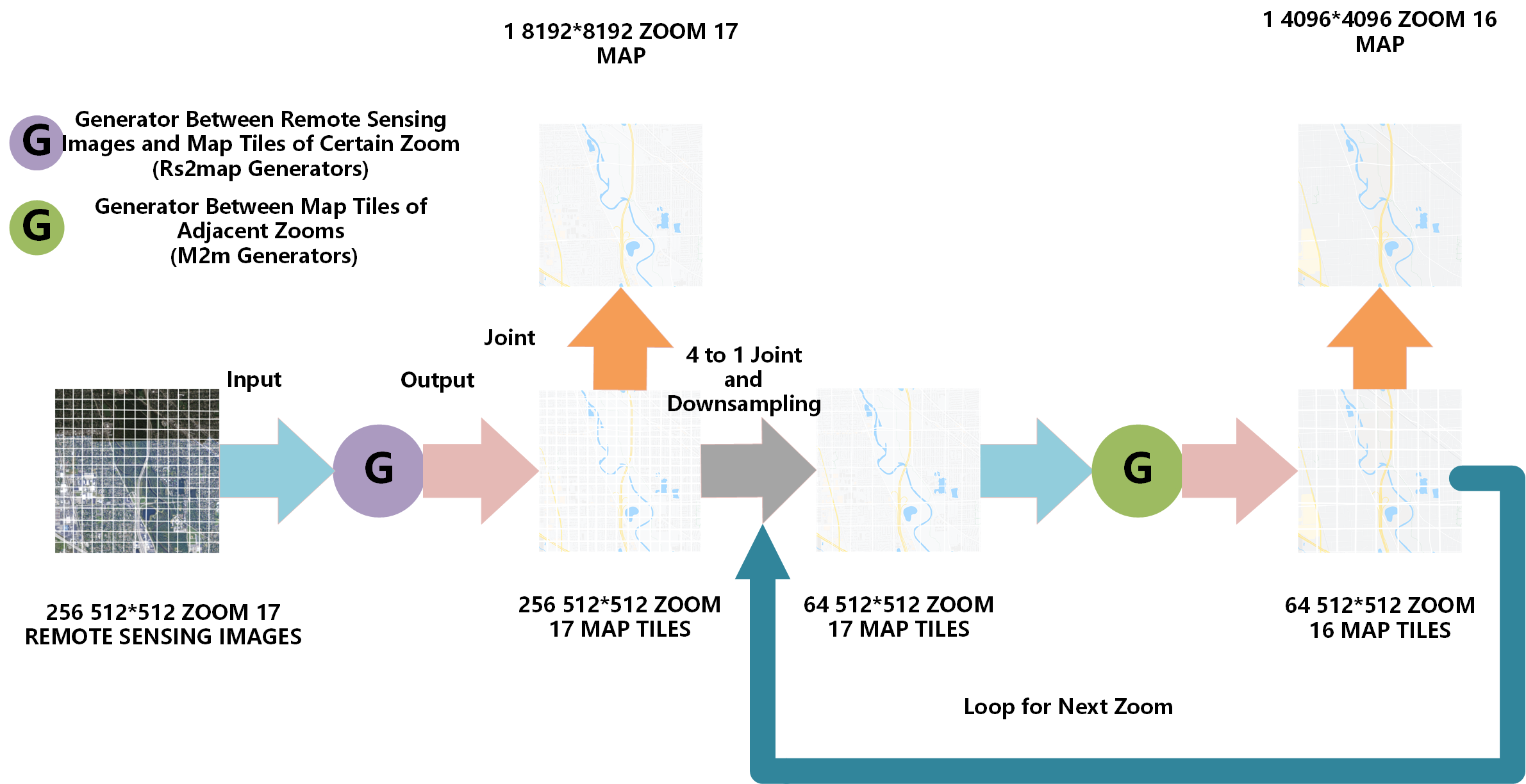}
	\caption{Architecture of the series strategy employed in this study.}
	\label{Figure 8}
\end{figure}

Fig. \ref{Figure 8} shows the architecture of the series strategy. The RSIs at a zoom level of 17 are translated to maps at a zoom level of 17 through an r2m generator at a zoom level of 17. These map tiles were merged to the zoom-17 map. Furthermore, four adjacent tiles were merged and downsampled to a 512 × 512 tile at a zoom level of 17 as the input for the next m2m translator. The above process is looped for each m2m translator to generate multi-scale maps.

The r2m translator is trained with paired samples of RSIs and maps at a zoom level of 17. The m2m translators are trained with paired samples of generated maps at a zoom level of $n$ and real maps at a zoom level of $n-1$ (\textbf{MM-pair}).

\subsection{Baseline: The Parallel Strategy}
\begin{figure}[htbp]
	\centering
	\includegraphics[scale=0.1]{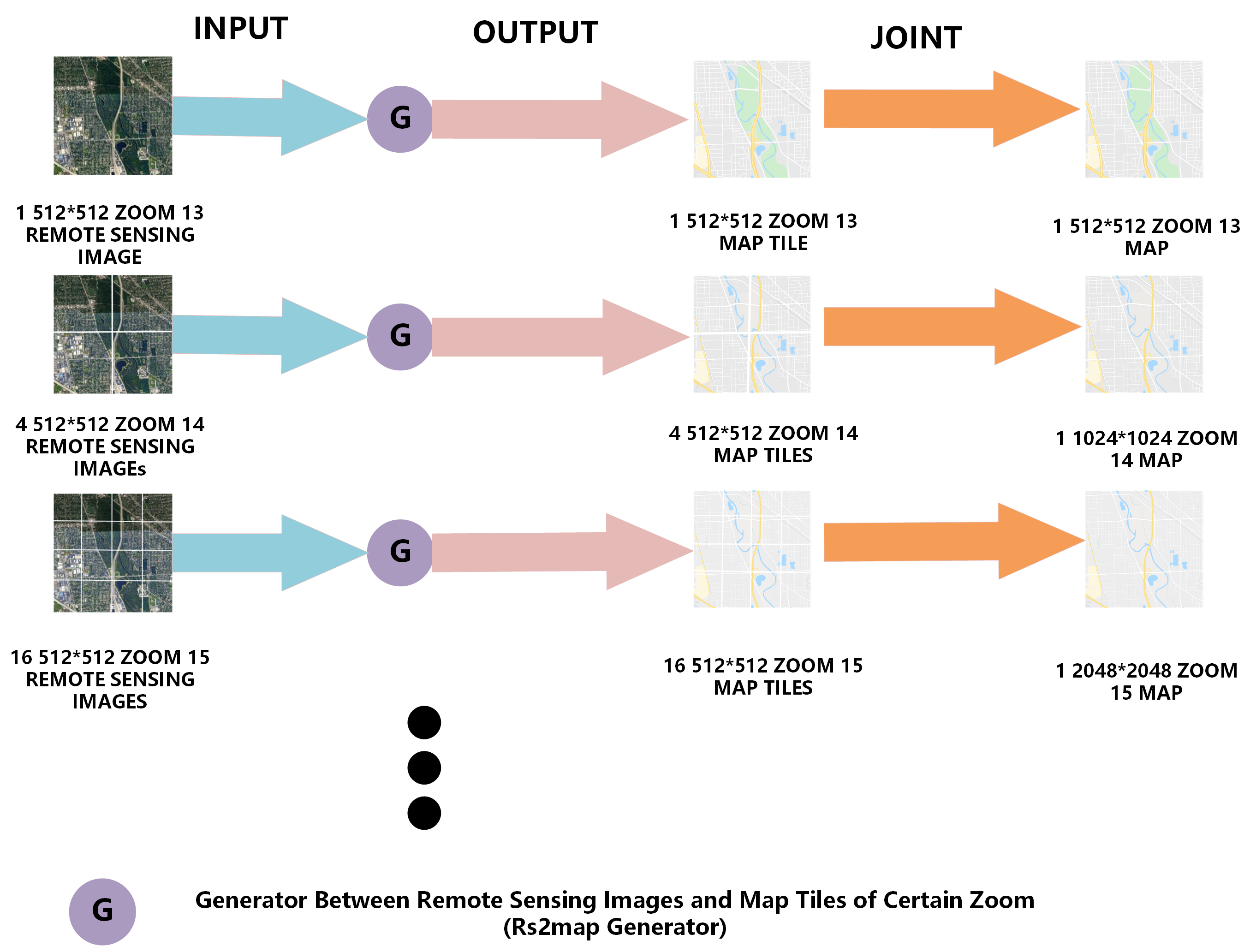}
	\caption{Architecture of the parallel strategy.}
	\label{Figure 5}
\end{figure}

Figure \ref{Figure 5} shows the architecture of the parallel strategy. Parallel scale-wise r2m generators translate RSIs of a certain zoom to maps of the same zoom respectively. Then outputted map tiles of each translator are jointed to the map of this zoom. Each r2m generator is trained with paired samples of RSIs and maps of corresponding zoom (\textbf{RM-pair}).

\section{Experiment and Results}

\begin{table}[htbp]\scriptsize
	\centering
	\caption{Capacity of Training and Testing Datasets}
	\begin{tabular}{cccc}
		\toprule[1pt]
		\multirow{2}[1]{*}{\textbf{SCALE}}&\textbf{TRAIN}  &\textbf{TEST}&	\multirow{2}[1]{*}{\textbf{PAIR TYPE}} \\
		&\textbf{11 Cities}&\textbf{Mexico City \& Tokyo}&\\
		\midrule[0.1pt]
		ZOOM 17&1086&1024&RM-pair\\
		\midrule[0.1pt]
		\multirow{2}[1]{*}{ZOOM 16}&1086&256&RM-pair\\
		&1086&0&MM-pair\\
		\midrule[0.1pt]
			\multirow{2}[1]{*}{ZOOM 15}&1086&64&RM-pair\\
		&1086&0&MM-pair\\
		\midrule[0.1pt]
			\multirow{2}[1]{*}{ZOOM 14}&773&16&RM-pair\\
		&876&0&MM-pair\\
		\midrule[0.1pt]
			\multirow{2}[1]{*}{ZOOM 13}&773&4&RM-pair\\
		&773&0&MM-pair\\
		\bottomrule[1pt]
	\end{tabular}%
	\label{table3}%
\end{table}%

\subsection{Datasets Construction}
The datasets include RSIs and corresponding map tiles collected from Google Maps covering the urban areas of 13 cities in North America, Europe, and Asia. Their zoom levels are from 13 to 17 and the spatial resolutions are 0.6–9.5 m/pixel. We neglect zoom levels higher than 17 because objects no longer maintain relative integrity on the maps, and neglect zoom levels lower than 13 because insufficient reasonable samples are accessible at these zooms. Table \ref{table3} presents the number and division of training and testing datasets.

\subsection{Metrics}
We adopted SSIM and ESSI into our metrics, and neglected metrics of GAN quality such as the inception score (IS) \cite{salimans_improved_2016} and Frechet inception distance (FID) \cite{heusel_gans_2017} because the experiment involves an assessment reference (real Google Map tiles); thus, full-reference image quality assessment indexes are more suitable for rs2map translation \cite{chen_smapgan_2020}.

Moreover, we applied IOU metric as it has been widely used in geoscience research for object detection \cite{9329036}, road extraction \cite{Zhou_2018_CVPR_Workshops}, classification \cite{9361068}, and cartography generalization \cite{ijgi9050338}. In brief, the IOU can measure whether pixels of objects on the RSI have been translated to pixels of corresponding objects on the generated map. As the colors of a map can indicate labels of various objects, we labeled objects as road and water on real maps through a clustering algorithm and measured the IOU of different categories of objects on the generated maps.

\subsection{Results}
\begin{table}[htbp]\tiny
	\centering
	\caption{Comparison of quantitative Results between parallel and series strategy}
		\begin{tabular}{cccccc}
		\toprule[1pt]
		\textbf{ZOOM}  & \textbf{STRATEGY} & \textbf{SSIM } & \textbf{ESSI } & \textbf{ROAD IOU} &\textbf{WATER IOU}\\
		\midrule[0.1pt]
		\multirow{2}[1]{*}{\textbf{17}} & Series & \multirow{2}[1]{*}{0.6164} & \multirow{2}[1]{*}{0.0573} & \multirow{2}[1]{*}{0.3244}&\multirow{2}[1]{*}{0.2200} \\
		& Parallel &       &       &  &\\
		\midrule[0.1pt]
		\multirow{2}[0]{*}{\textbf{16}} & Series & \textbf{0.6731} & \textbf{0.1488} & \textbf{0.3329}&\textbf{0.2347} \\
		& Parallel & 0.5821 & 0.1142 & 0.2162&0.2125 \\
		\midrule[0.1pt]
		\multirow{2}[0]{*}{\textbf{15}} & Series & 0.5720 & \textbf{0.1438} & \textbf{0.3319}& \textbf{0.2999} \\
		& Parallel & \textbf{0.5884} & 0.1046 & 0.1836&0.2058 \\
		\midrule[0.1pt]
		\multirow{2}[0]{*}{\textbf{14}} & Series & \textbf{0.4605} & \textbf{0.1056} & \textbf{0.3132}&\textbf{0.3812} \\
		& Parallel & 0.4037 & 0.0802 & 0.1732&0.1507 \\
		\midrule[0.1pt]
		\multirow{2}[1]{*}{\textbf{13}} & Series & \textbf{0.4941} & \textbf{0.1778} & \textbf{0.1732}&\textbf{0.4745} \\
		& Parallel & 0.4123 & 0.08245 & 0.1632&0.2633 \\
		\midrule[0.1pt]
		\textbf{AVERAGE}&&\multirow{2}[1]{*}{\textbf{11.69\%}}&\multirow{2}[1]{*}{\textbf{53.78\%}}&\multirow{2}[1]{*}{\textbf{55.42\%}}&\multirow{2}[1]{*}{\textbf{72.34\%}}\\
		\textbf{INCREASE}&&&&&\\
		\bottomrule[1pt]
	\end{tabular}%
	\label{table2}%
\end{table}%
\begin{figure}[htbp]
	\centering
	\includegraphics[scale=0.25]{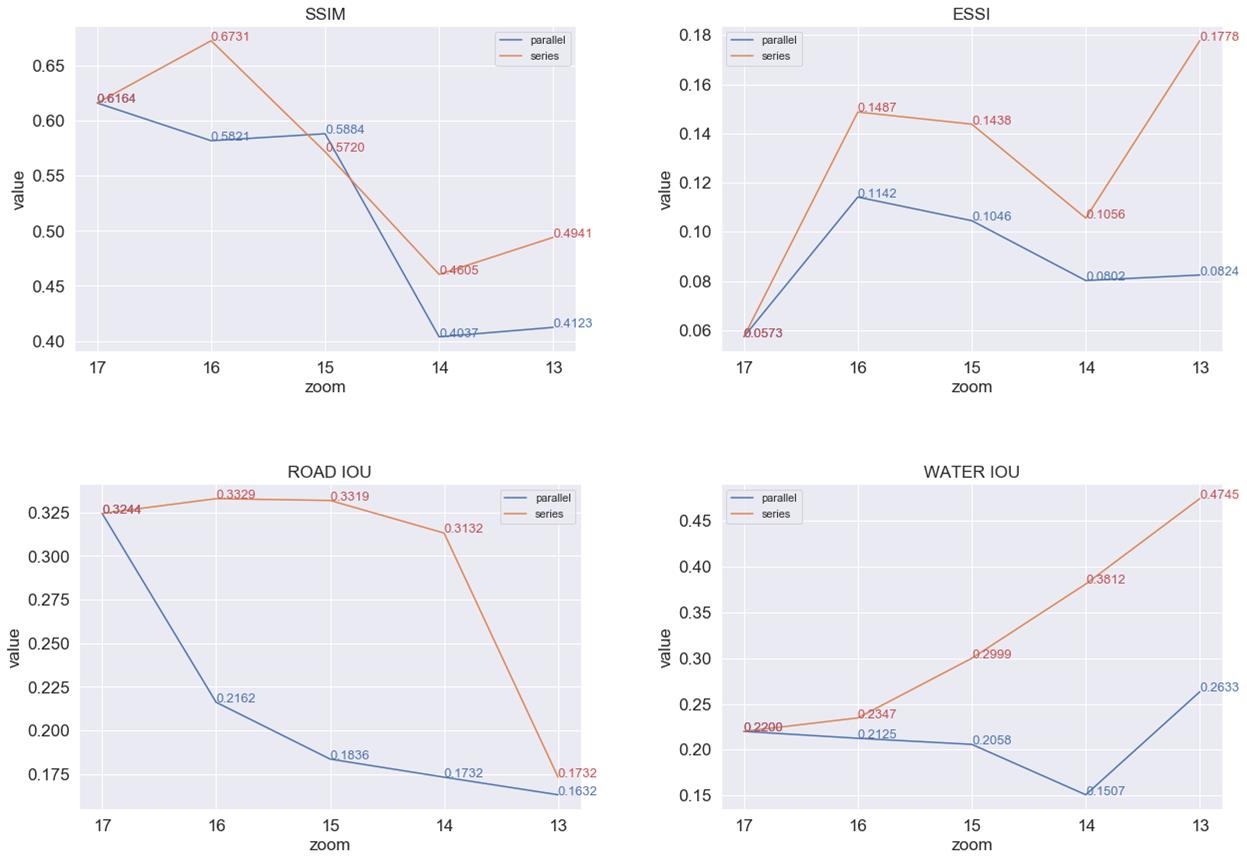}
	\caption{Variation trends of the evaluation metrics employed in this study. SSIM: structural similarity index; ESSI: edge structural similarity index; ROAD IOU: intersection over union (road); WATER IOU: intersection over union (water). }
	\label{Figure 7}
\end{figure}
Table \ref{table2} present the quantitative results of SSIM, ESSI, and IOUs (road and water), and Fig. \ref{Figure 7} shows the trend of metrics with decreasing zoom level. Generally, the series strategy results in the best metrics in the experiments, especially for the ESSI and water IOU. Regarding the road IOU, the series strategy generates stable results from a zoom level of 17 to 14, whereas the parallel strategy results exhibit a consistent decrease. At a zoom level of 13, the road IOU of the series strategy declines sharply, which may result from the elimination of many roads at this zoom level.  
\begin{figure}[htbp]
	\centering
	\includegraphics[scale=0.12]{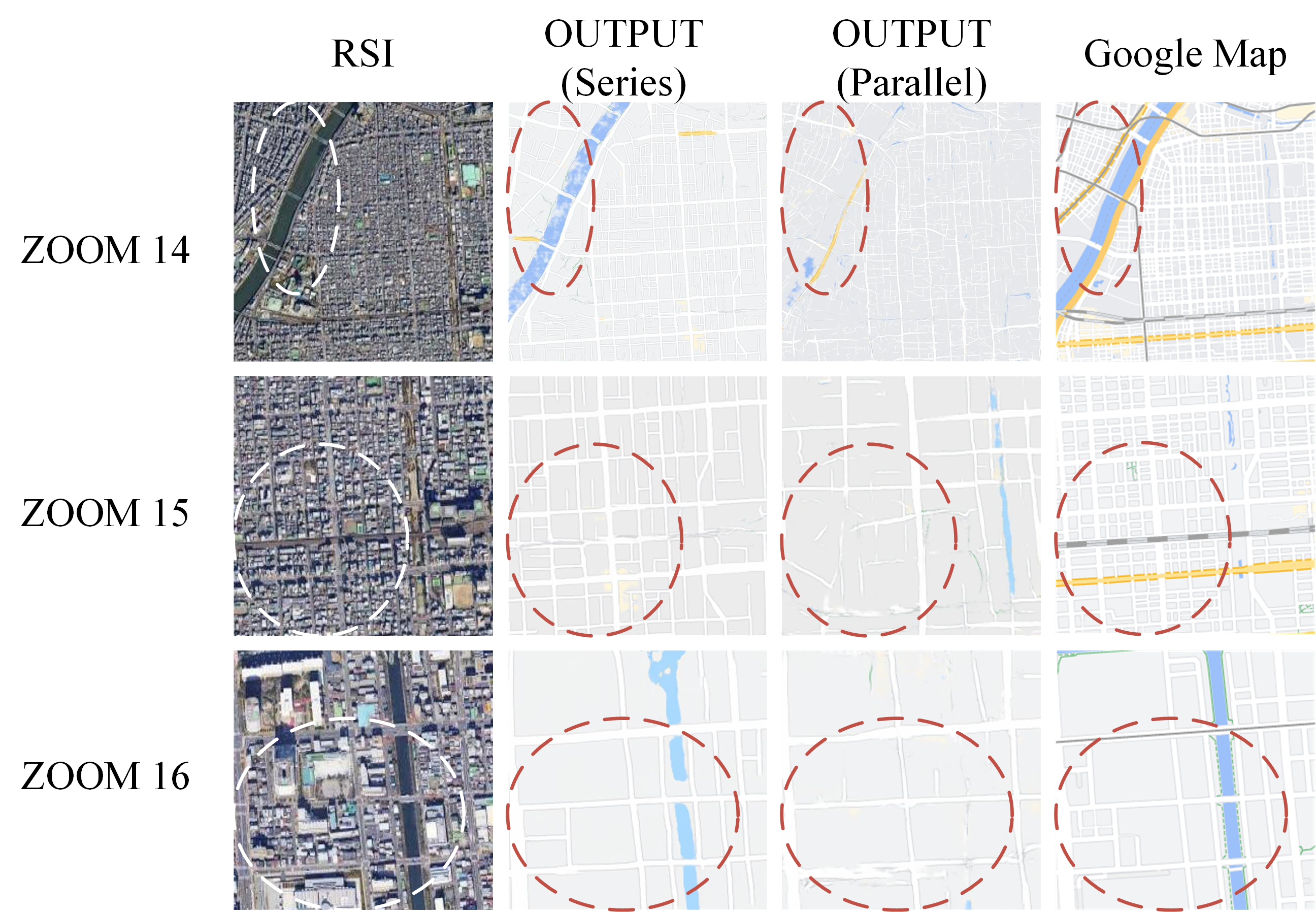}
	\caption{Examples of multi-scale maps generated by the series and parallel strategies at different zoom levels.}
	\label{Figure3}
\end{figure}

Fig. \ref{Figure3} shows the qualitative results, which reveal better details and road continuity of outputs in the series strategy than in the parallel strategy. Both quantitative and qualitative results prove that maps generated by the series strategy exhibit better quality than those generated by the parallel strategy, which proves the effectiveness of the series strategy design.

\section{Conclusion and Future Work}
In this study, we analyzed limitation of the parallel strategy: RS-m inconsistency and multi cross-domain translation. Then we improved the quality of multi-scale map generation by replacing parallel generators with series generators. First, this modified strategy uses high-resolution RSIs of the largest scale as inputs instead of multi-scale RSIs, which avoids RS-m inconsistency. Second, the distribution gap between maps of neighboring scales is smaller than that between RSIs and maps of the same scale according to the EMD. The series strategy replaces cross-domain translation with near-intra-domain translation, with lower computation costs for generators. Our experimental results revealed that maps generated using the series strategy exhibit quantitatively and qualitatively better quality, which proves the effectiveness of our analysis and design.

Further studies on multi-scale rs2map translation are required. The experiments in this study were only conducted from zooms levels of 17 to 13; thus, the performance of the series strategy at lower zoom levels remains to be determined. A combined series and parallel strategy should also be studied. Furthermore, Pix2PixHD consumes a lot of memory that we can only train generators separately owing to equipment limitations. Therefore, this strategy could be improved through the global training of generators.

\section*{Acknowledgement}
Thanks to dear Kiki and the keyboard from her typing this article. Thanks to Editage (www.editage.cn) for English language editing. 

\ifCLASSOPTIONcaptionsoff
  \newpage
\fi



\bibliographystyle{IEEEtran}
\bibliography{refs.bib}
\end{document}